\pdfoutput=1

\documentclass[11pt]{article}

\usepackage[final]{ACL2023}

\usepackage{times}
\usepackage{latexsym}
\usepackage{footmisc}

\usepackage[T1]{fontenc}

\usepackage[utf8]{inputenc}
\usepackage{graphicx}
\usepackage{microtype}

\usepackage{inconsolata}
\usepackage{booktabs} 
\usepackage{multirow}

\usepackage{graphicx}
\usepackage{amsmath,amsfonts,bm}

\def\eqref#1{equation~\ref{#1}}

\def\1{\bm{1}}

\DeclareMathAlphabet{\mathsfit}{\encodingdefault}{\sfdefault}{m}{sl}
\SetMathAlphabet{\mathsfit}{bold}{\encodingdefault}{\sfdefault}{bx}{n}

\renewcommand{\paragraph}[1]{\medskip\noindent\textbf{#1.~}}

\newcommand{\MName}{Kwai-STaR}

\newcommand{\blfootnote}[1]{%
  \begingroup
  \renewcommand\thefootnote{}\footnote{#1}%
  \addtocounter{footnote}{-1}%
  \endgroup
}

\renewcommand{\thefootnote}{\fnsymbol{footnote}}
\title{\MName{}: 
Transform LLMs into State-Transition Reasoners
}

\author{
    Xingyu Lu\textsuperscript{1,2}$^{\dagger}$,
    Yuhang Hu\textsuperscript{2,3}$^{\dagger}$,
    Changyi Liu\textsuperscript{2},
    {Tianke Zhang\textsuperscript{2}},
    {\bf Zhenyu Yang\textsuperscript{2,3}},\\
    {\bf Zhixiang Ding\textsuperscript{2,3}},
    {\bf Shengsheng Qian\textsuperscript{3}},
    {\bf Meng Du\textsuperscript{2}},
    {\bf Ruiwen Kang\textsuperscript{2}},
    {\bf Kaiyu Tang\textsuperscript{2}},\\
    {\bf Fan Yang\textsuperscript{2}},
    {\bf Tingting Gao\textsuperscript{2}},
    {\bf Di Zhang\textsuperscript{2}},
    {\bf Hai-Tao Zheng\textsuperscript{1,4}$^{\ast}$},
    {\bf Bin Wen\textsuperscript{2}$^{\ddagger}$\thanks{\ \ Corresponding authors: Hai-Tao Zheng (\url{zheng.haitao@sz.tsinghua.edu.cn}) and Bin Wen (\url{wenbin@kuaishou.com})}}\\
    \textsuperscript{1}Shenzhen International Graduate School, Tsinghua University \\
    \textsuperscript{2}Kuaishou Technology \\
    \textsuperscript{3}Institute of Automation, Chinese Academy of Sciences\\
    \textsuperscript{4} Pengcheng Laboratory, Shenzhen, China, 518055 \\
}
\begin{document}
\maketitle

\begin{abstract}
Mathematical reasoning presents a significant challenge to the cognitive capabilities of LLMs.
Various methods have been proposed to enhance the mathematical ability of LLMs. However, few recognize the value of \textbf{state transition} for LLM reasoning. 
In this work, we define mathematical problem-solving as a process of transiting from an initial unsolved state to the final resolved state, and propose \textbf{\MName{}} framework, which transforms LLMs into \textbf{S}tate-\textbf{T}r\textbf{a}nsition \textbf{R}easoners to improve their intuitive reasoning capabilities. Our approach comprises three main steps: 
(1) Define the state space tailored to the mathematical reasoning. (2) Generate state-transition data based on the state space. 
(3) Convert original LLMs into State-Transition Reasoners via a curricular training strategy. 
Our experiments validate the effectiveness of \MName{} in enhancing mathematical reasoning: After training on the small-scale \MName{} dataset, general LLMs, including Mistral-7B and LLaMA-3, achieve considerable performance gain on the GSM8K and GSM-Hard dataset. Additionally, the state transition-based design endows \MName{} with remarkable training and inference efficiency.
Further experiments are underway to establish the generality of \MName{}. 
\blfootnote{$\dagger$: Equal contribution.}

\blfootnote{$\ddagger$: Project lead}
\end{abstract}
\section{Introduction}

As an intricate problem for artificial intelligence\cite{survey1}, mathematical reasoning poses a significant challenge to the cognitive abilities of large language models (LLMs).
Recently, various efforts have been dedicated to enhancing LLMs' mathematical capabilities, achieving significant progress.

A range of work focuses on enhancing the inherent mathematical abilities of LLMs through training:
GPT \cite{gpt4} and LLaMA \cite{llama3} incorporate massive mathematical corpus to improve general intelligence, OpenMath \cite{openmath} and MetaMath \cite{metamath} specialize general LLMs into math-tailored models with large-scale in-domain datasets. For these methods, organizing mathematical corpus into structured data instances becomes an essential step. 
Another line of research, including CoT \cite{cot} and Best-of-N \cite{bon},  explores how to fully harness the potential of LLMs during inference to boost mathematical performance. Recently, Monte Carlo Tree Search \cite{rstar,berry} and Process Reward Model \cite{shepherd} have been applied; they achieve remarkable results by decomposing problem-solving process into multiple steps and providing intermediate rewards. 

Whether for training or inference paradigm, the modeling of mathematical reasoning process is playing an increasingly significant role, which determines how to appropriately organize the corpus and decompose the mathematical problem. In this paper, we provide a novel perspective to organize the reasoning process of LLMs: the solution of a mathematical problem can be intuitively viewed as an ordered sequence of transition from an initial unsolved state (\textit{original question}) to a final resolved state (\textit{correct answer}). 

Based on this observation, we manage to design a state transition paradigm to assist mathematical reasoning of LLM. We present \MName{}, a framework to transform general LLMs into state-transition reasoners, which systematically solve problems by performing state transition. 

\MName{} takes three steps to incorporate state transition with LLM reasoning: (1) \textbf{State space definition}: We develop an action set and constrain the LLM to solve problems as a State-Transition Reasoner (STaR): The LLM takes one action from the action set at a time to transition the current state into a new state.  (2) \textbf{State-transition data construction}. Although advanced LLMs possess remarkable instruction-following capabilities, we still take appropriate training to help general language models master state-by-state reasoning. Hence, we construct a small-scale  state-transition dataset with meticulous generation instructions and pipeline.
(3) \textbf{Curricular Training Strategy}. Our dataset contains two types of instances: a majority of correct cases and a minority of wrong-then-verified cases from the data generator and trained reasoner.
To maximize learning efficiency, our training strategy consists of two stages: a fundamental stage and an advanced stage. The former trains the model with the majority right cases, enabling it to grasp the state-transition manner and to solve relatively simple problems. The latter leverages pairs of wrong and verified cases to further strengthen the model’s proficiency.

Using the above method, we construct a small-scale state-transition dataset from the GSM8K training data, our dataset contains 20K right instances and approximately 3K wrong-then-verified instances. We test \MName{}'s effect on various general LLMs. Our experiment indicates that \MName{} framework significantly improves the performance of all tested general LLMs on two benchmarks. Moreover, compared to other data augmentation methods, the state-transition-based \MName{} dataset offers higher data efficiency. In contrast to inference-time enhancement methods, \MName{} achieves single-pass accuracy comparable to those methods' multi-pass accuracy, without complex inference paradigm.

We summarize our contributions as follows:

\begin{itemize}
    \item We provide the novel perspective of state transition to model the mathematical reasoning of LLMs and construct a state-transition dataset.
    
    \item We propose \MName{} framework to enhance LLM reasoning through state transitions. \MName{} effectively transforms models of various scales into STaRs, significantly improving their mathematical performance.
    
    \item \MName{} achieves remarkable performance and efficiency, revealing the great potential of state-space strategies in enhancing LLM reasoning. We are actively extending our state-space strategies to broader scenarios.
\end{itemize}

\section{The \MName{} Framework}
\label{sec:Method}
In the following sections, we first introduce the relevant concepts of the State-Transition Reasoner, then describe three steps of \MName{} to transform LLMs into state-transition reasoner: 
1) Define the mathematical state space.
2) Construct high-quality state-transition data.
3) Teach LLMs to solve problems by state transition.

\subsection{Relevant Concepts}
We adopt RL concepts to formalize the mathematical problem solving as a state transition process.

\begin{itemize}
    \item \textbf{State}: A specific point in the problem-solving process ranging from the \textbf{initial} state (\textit{original question}) to the \textbf{final} state (\textit{correct answer}).
    
    \item \textbf{State Transition Reasoner}: A LLM that systematically solves problems by executing actions in the action set to transition from the initial state to the final state.
    
    \item \textbf{Action}: Operations performed by the reasoner to transit states. \MName{}'s action set includes 7 kinds of actions (Table \ref{tab:action_set}).

\end{itemize}

\begin{table}[h]
\small
\scalebox{0.95}{
\centering
    \begin{tabular}{lp{5cm}}
        \toprule
        \textbf{Action} & \textbf{Definition} \\
        \toprule
        \textbf{Formalize} & {\small Formalize the question mathematically} \\
        \textbf{Decompose} & {\small Divide the original question.} \\
        \textbf{Solve Subques} & {\small Provide solution for one subquestion.} \\
        \textbf{Solve Parent} & {\small Solve the original question.} \\
        \textbf{Verify} & {\small Check the correctness of current state.} \\
        \textbf{Backtrack} & {\small Backtrack to the last correct state.} \\
        \textbf{Summarize} & {\small State the final answer.} \\
        \toprule
    \end{tabular}
}
\vskip -0.1in
\caption{Action Set of \MName{}.}
\vskip -0.2in
\label{tab:action_set}
\end{table}

\subsection{\MName{} State Space}

The design of \MName{}'s state space is guided by the divide-and-conquer principle. First, the reasoner formalizes the original problem into mathematical expressions, listing the variables and their relationships. Then, the reasoner decomposes the complex problem into multiple subquestions and solves them individually. Finally, the reasoner combines subquestions' answers to solve the original problem. If an incorrect intermediate result is generated, the reasoner is expected to perform a Verify action and backtrack to the last correct state. 

Through the above definitions, the reasoning process of the LLM is formalized as a series of state transitions within the state space. 

\begin{figure*}[!hbtp]
    \centering
    \includegraphics[width=0.95\linewidth]{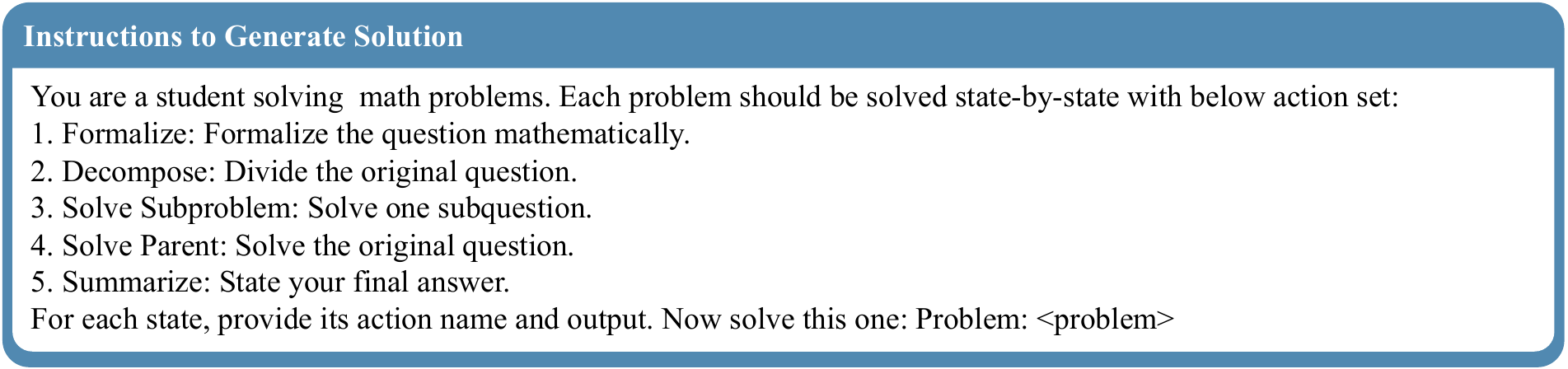}
    \vskip -0.1in
    \caption{\footnotesize{Example of instruction for state-transition data construction.}}
    \label{fig:construct_prompt}
    \vskip -0.1in
\end{figure*}
\subsection{State-Transition Data Construction}

Although state-of-art LLMs can follow instructions to perform state transition, the ability of smaller and weaker models may be not guaranteed. Thus, to improve reasoning ability of various LLMs broadly, we construct  state-transition reasoning data with small scale and high quality (Table \ref{tab:data_comp}) to teach LLMs how to perform state transition.

\begin{table}[!ht]
\centering

\vskip -0.07in
\scalebox{0.85}{
\begin{tabular}{lcc}
    \toprule
    \textbf{Dataset} & \textbf{Format } &\textbf{Scale} \\
    \midrule
    MetamathQA & Rephrased QA pairs & 395 K \\
    MathGenie & Rephrased QA pairs & 284 K \\
    \midrule
    \MName{} & State-Transition data & 20 K \\
    \midrule
\end{tabular}
}
\vskip -0.1in
\caption{\footnotesize Data format and scale of different math dataset.}
\vskip -0.1in
\label{tab:data_comp}
\end{table}
We construct \MName{} dataset with training sets of several math benchmarks. The data construction process of \MName{} involves two stages: In \textbf{Stage \uppercase\expandafter{\romannumeral 1}}, the action list in the instructions excludes \textit{Verify} and \textit{Backtrack}, the data generator act as students to solve problems with the provided actions. In \textbf{Stage \uppercase\expandafter{\romannumeral 2}}, for the wrong cases from the data-generator and fine-tuned models on train set, we supply reference answers and instruct the data generator to correct the erroneous cases as a teacher, using the complete action list. The state transition paths in incorrect and verified cases naturally form accept-reject pairs, which serve as reinforcement learning samples in the \textbf{Advanced Refinement} stage of training.

\begin{figure*}[!hbtp]
    \centering
    \includegraphics[width=0.95\linewidth]{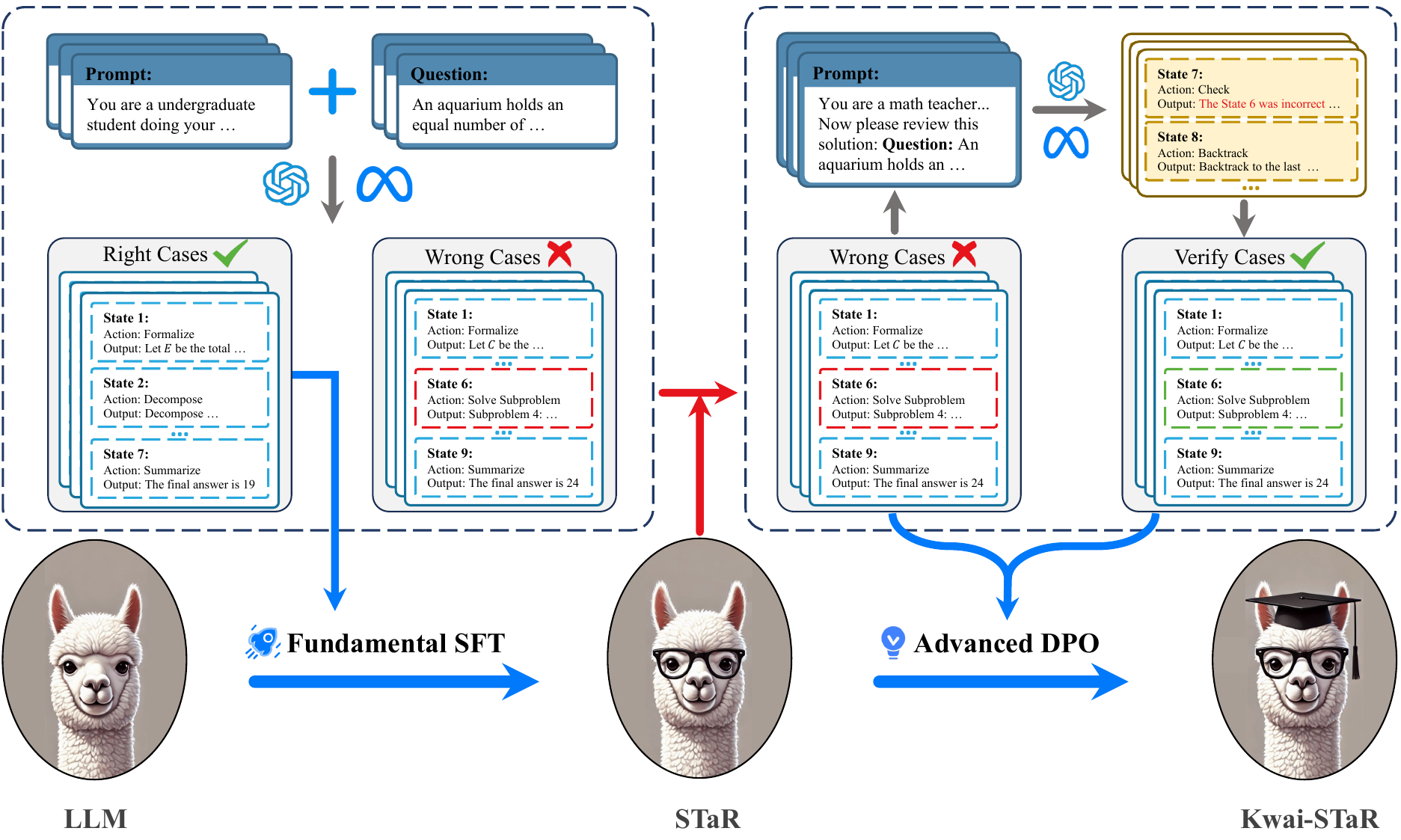}
    \vskip -0.1in
    \caption{\footnotesize{The overall process of \MName{}'s data construction and model training. For data construction (above), we prompt the data generators with pre-defined state-space instruction  to construct state-transition data. For model training (below), we employ a two-stage approach: in the first stage, we perform supervised fine-tuning (SFT) on right-case data; in the second stage, we apply DPO training using wrong-then-verified cases.}}
    \label{fig:overview}
    \vskip -0.1in
\end{figure*}

\subsection{Transform LLMs into State-Transition Reasoner}
Corresponding to the data construction process, \MName{}'s training process also consists of two progressive stages: \textbf{Fundamental Training} and \textbf{Advanced Refinement}, each stage uses the data from the respective data construction stage for training. \MName{}'s training strategy can also be viewed as a state-transition process: As shown in Figure \ref{fig:overview}, the \textbf{Fundamental Training} stage first transforms standard LLMs into basic State-Transition Reasoners, then the \textbf{Advanced Refinement} stage further equip the basic reasoner with superior mathematical proficiency.

In the \textbf{Fundamental Training} stage, the correct cases from Stage \uppercase\expandafter{\romannumeral 1} are adopted to train the model with commonly used next-token prediction loss:
\vskip -0.1in
\begin{equation*}
    \mathcal{L}_{\text{NTP}} = -\sum_{t=1}^{T} \log P(y_t \mid y_{<t}; \theta)
\end{equation*}
Here’s the translation:

The correct cases constitute the majority of the training set and are relatively less challenging, hence they are assigned to help the model master the process of solving mathematical problems through state transitions and learning solution methods for most problems.

In the \textbf{Advanced Refinement} stage, we take accept-reject pairs from more challenging wrong cases as data for DPO training. 

\vskip -0.3in
\begin{multline*}
\mathcal{L}_{\mathrm{DPO}}(\pi_{\theta}; \pi_{\mathrm{ref}}) 
= -\mathbb{E}_{(x, y_a, y_r) \sim \mathcal{D}} \Bigg[ \log \sigma \Bigg( 
 \\
 \beta\log \left( \frac{\pi_{\theta}(y_a \mid x)}{\pi_{\mathrm{ref}}(y_a \mid x)} \right) 
- \beta \log \left( \frac{\pi_{\theta}(y_r \mid x)}{\pi_{\mathrm{ref}}(y_r \mid x)} \right) \Bigg) \Bigg]
\end{multline*}

In this formula, $\pi_{\text{ref}}$ refers to the language model trained in the fundamental stage, $\pi_{\theta}$ denotes the language model updated through DPO, $y_a$ represents the verified accepted solution, $y_r$ is the rejected wrong solution, and $x$ stands for the input problem.

Since the wrong cases stem from the mistakes made by the generator and the trained model, they are challenging enough to teach the model how to tackle hard problems and further addressing the shortcomings left by \textbf{Fundamental Training}.

After the two-stage training, the model acquires outstanding capability to solve problems by state transitions. During inference, the model can explore the state space, reasoning from the input problem to achieve the final answer.

\section{Experiments}
\label{sec:Exp}
\subsection{Experiment Setup}
\textbf{Datasets.}
We choose GSM8K benchmark for experiment. On its training split, we adopt advanced language models including LLaMA-3.1-Insturct-70B and GPT-4o as generator to construct Kwai-STaR dataset, while strictly limiting the data scope. 

\textbf{Models.} We employ state-of-the-art LLMs including Mistral-7B\cite{mistral}, LLaMA3-Instruct series, Phi3-mini-4k\cite{phi3} as base models, ranging from 4B to 8B.

\textbf{Compared methods.}
We select three types of powerful methods as baselines:
(1) \textbf{Training refinement}: including MetaMathQA.
(2) \textbf{Inference enhancement}: such as Self-Consistency and LLaMA-Berry.
(3) \textbf{Regular methods}: including CoT, few-shot and SFT.

\textbf{Details.}
For methods requires training, we follow their reported configurations. We run all experiments with LLaMA-Factory framework on same device for consistency. To maintain same parameter count, our experiment does not include reward models in some methods. For \MName{}, all models are fine-tuned with LoRA for efficiency. 

\begin{table}[h]
\centering
\small
\scalebox{0.95}{
    \begin{tabular}{l|p{2.2cm}}
        \textbf{Hyperparameter} & \textbf{Value} \\
    \hline
        \textbf{lora\_rank} & 16 \\
        \textbf{learning\_rate} & 1.0 e-4 \\
        \textbf{epochs} & 10 \\
        \textbf{optimizer} & AdamW \\
        \textbf{lr scheduler} & cosine decay \\
        \textbf{batch\_size} & 32 \\
    \end{tabular}
}
\vskip -0.1in
\caption{Hyperparameter Settings.}
\vskip -0.2in
\label{tab:hyperparams}
\end{table}

\begin{table*}[!ht]
\centering
\scalebox{0.95}{
    \begin{tabular}{clcccc}
    \toprule
    \textbf{Dataset} & \textbf{Method} & \textbf{Mistral-7B} & \textbf{LLaMA3-8B} & \textbf{LLaMA3.1-8B} & \textbf{Phi3-mini-4k} \\
    \midrule
    \multirow{10}{*}{GSM8K}  & 0-shot CoT         & 17.89  & 68.38  & 76.60  & 20.17  \\
                            & 8-shot CoT         & 36.46  & 74.53  & --  & 83.45  \\
                            & SFT                & 68.16  & 75.36  & 79.45  & 81.13  \\
                            & MetaMathQA         & \underline{72.55}  & 79.08  & 81.50  & --  \\
                            & SC(maj@128)          & 57.25  & 84.69  & 88.93  & 88.68  \\
                            & R-STaR(maj@1)        & 53.30  & 83.85 & --  & 85.97  \\
                            & R-STaR(maj@8)        & 62.17  & 86.05 & --  & \underline{90.45}  \\
                            & LLaMA-Berry(maj@1)   & --  & {68.40}  & 76.60  & --  \\
                            & LLaMA-Berry(maj@8)& --  & \underline{86.40}  & \underline{89.80}  & --  \\
                            & \textbf{\MName{} (maj@1)} & \textbf{80.52}  & \textbf{86.81}  & \textbf{90.22}  & \textbf{90.52}  \\
    \bottomrule
    \multirow{7}{*}{GSM-Hard} & 0-shot CoT       & 5.16  & 14.94  & --  & 33.73  \\
                              & 8-shot CoT       & 13.57  & 25.63  & --  & 40.63  \\
                              & SC(maj@128)        & 25.01  & \textbf{31.16}  & --  & 45.56  \\
                              & R-STaR(maj@8)      & \underline{27.45}  & {30.93} & --  & \underline{45.79}  \\
                              & LLaMA-Berry(maj@1) & --  & 14.90  & {32.75}  & --  \\
                              & LLaMA-Berry(maj@8) & --  & 30.20  & \textbf{35.78}  & --  \\
                              & \textbf{\MName{}(maj@1)} & \textbf{30.86}  & \underline{31.01}  & \underline{35.18}  & \textbf{48.52}  \\
    \bottomrule
    \end{tabular}
}
\caption{Performance comparison between \MName{} and other methods across various base models (Two LLaMA models are \textbf{Instruct} version). We test them on the GSM8K and GSM-Hard benchmark. The best results are in \textbf{bold}.}
\label{table:experiment_results}
\end{table*}
\subsection{Primary Results}
In Table \ref{table:experiment_results}, we present the accuracy of \MName{} compared to other baselines on the GSM8K and GSM-Hard Benchmark.
As can be seen, four \MName{} models achieves the highest accuracy on the GSM8K dataset. On the GSM-Hard dataset, \MName{} also demonstrates either optimal or sub-optimal performance. Since the current training data only includes the training set of GSM8K, the success on GSM-Hard demonstrates that \MName{}'s enhancement of LLM reasoning can adapt to more complex mathematical scenarios.

Beyond performance gain, \MName{} also offers efficiency advantages compared to other baselines:
(1) \textbf{Training}: Compared with data-augmentation methods such as MetaMATH, Kwai-STaR achieves greater performance improvements with a smaller data scale and fewer trainable parameters. We attribute this to the high quality and structured format of \MName{}'s data, demonstrating the feasibility of state transitions in mathematical problem-solving.
(2) \textbf{Inference}: Compared to inference-time enhancement methods like CoT, Self-Consistency, and MCTS-based approaches, \MName{}’s majority@1 accuracy is comparable to their majority@n accuracy, yet without complex inference paradigms and expensive inference cost.
These advantages imply that \MName{} effectively enhances the LLMs' intuitive reasoning ability with minimal training costs, we are convinced that the significant advantages of \MName{} demonstrate the feasibility and great potential of state-transition paradigms for enhancing the reasoning capabilities of LLMs. We are managing to provide more experiment results and analysis to further show \MName{}'s ability.
\section{Conclusion and Future Work}
In this paper, we introduce the \MName{} approach: By defining the state space, constructing state-transition data, and applying progressive training strategy, we transform a general LLM into a state-transition reasoner, thereby enhancing its reasoning capabilities for tackling mathematical problems.
Our experiments show that \MName{} significantly enhances LLMS' mathematical performance, demonstrating the crucial role and potential of state spaces in supporting LLM reasoning.

Due to time constraints, currently we have not finished testing \MName{}'s feasibility in other scenarios. We are actively working on it to provide additional experimental results in more diverse and general settings to further demonstrate the generalizability of the \MName{} approach.

\section{Limitations}
\label{sec:Limitations}
The current Kwai-STaR has only validated its effectiveness in the field of mathematics. While the mathematical domain is both challenging and representative, the potential of state-space for enhancing LLM reasoning in general scenarios remains unverified, which limits the generalizability of the Kwai-STaR. We are working on releasing updated versions of Kwai-STaR to address this limitation. Currently, the design of the state space is primarily manual. Although this approach has yielded good results, it lacks completeness and automation, which are also directions for our future work. Currently, we lack a theoretical explanation of how state space improves LLM reasoning.

\bibliographystyle{acl_natbib}

\begin{thebibliography}{12}
\expandafter\ifx\csname natexlab\endcsname\relax\def\natexlab#1{#1}\fi

\bibitem[{Abdin et~al.(2024)Abdin, Aneja, Awadalla, Awadallah, Awan, Bach, Bahree, Bakhtiari, Bao, Behl et~al.}]{phi3}
Marah Abdin, Jyoti Aneja, Hany Awadalla, Ahmed Awadallah, Ammar~Ahmad Awan, Nguyen Bach, Amit Bahree, Arash Bakhtiari, Jianmin Bao, Harkirat Behl, et~al. 2024.
\newblock Phi-3 technical report: A highly capable language model locally on your phone.
\newblock \emph{arXiv preprint arXiv:2404.14219}.

\bibitem[{Achiam et~al.(2023)Achiam, Adler, Agarwal, Ahmad, Akkaya, Aleman, Almeida, Altenschmidt, Altman, Anadkat et~al.}]{gpt4}
Josh Achiam, Steven Adler, Sandhini Agarwal, Lama Ahmad, Ilge Akkaya, Florencia~Leoni Aleman, Diogo Almeida, Janko Altenschmidt, Sam Altman, Shyamal Anadkat, et~al. 2023.
\newblock Gpt-4 technical report.
\newblock \emph{arXiv preprint arXiv:2303.08774}.

\bibitem[{Ahn et~al.(2024)Ahn, Verma, Lou, Liu, Zhang, and Yin}]{survey1}
Janice Ahn, Rishu Verma, Renze Lou, Di~Liu, Rui Zhang, and Wenpeng Yin. 2024.
\newblock Large language models for mathematical reasoning: Progresses and challenges.
\newblock \emph{arXiv preprint arXiv:2402.00157}.

\bibitem[{Dubey et~al.(2024)Dubey, Jauhri, Pandey, Kadian, Al-Dahle, Letman, Mathur, Schelten, Yang, Fan et~al.}]{llama3}
Abhimanyu Dubey, Abhinav Jauhri, Abhinav Pandey, Abhishek Kadian, Ahmad Al-Dahle, Aiesha Letman, Akhil Mathur, Alan Schelten, Amy Yang, Angela Fan, et~al. 2024.
\newblock The llama 3 herd of models.
\newblock \emph{arXiv preprint arXiv:2407.21783}.

\bibitem[{Jiang et~al.(2023)Jiang, Sablayrolles, Mensch, Bamford, Chaplot, de~las Casas, Bressand, Lengyel, Lample, Saulnier et~al.}]{mistral}
AQ~Jiang, A~Sablayrolles, A~Mensch, C~Bamford, DS~Chaplot, D~de~las Casas, F~Bressand, G~Lengyel, G~Lample, L~Saulnier, et~al. 2023.
\newblock Mistral 7b (2023).
\newblock \emph{arXiv preprint arXiv:2310.06825}.

\bibitem[{Qi et~al.(2024)Qi, Ma, Xu, Zhang, Yang, and Yang}]{rstar}
Zhenting Qi, Mingyuan Ma, Jiahang Xu, Li~Lyna Zhang, Fan Yang, and Mao Yang. 2024.
\newblock Mutual reasoning makes smaller llms stronger problem-solvers.
\newblock \emph{arXiv preprint arXiv:2408.06195}.

\bibitem[{Stiennon et~al.(2020)Stiennon, Ouyang, Wu, Ziegler, Lowe, Voss, Radford, Amodei, and Christiano}]{bon}
Nisan Stiennon, Long Ouyang, Jeffrey Wu, Daniel Ziegler, Ryan Lowe, Chelsea Voss, Alec Radford, Dario Amodei, and Paul~F Christiano. 2020.
\newblock Learning to summarize with human feedback.
\newblock \emph{Advances in Neural Information Processing Systems}, 33:3008--3021.

\bibitem[{Toshniwal et~al.(2024)Toshniwal, Moshkov, Narenthiran, Gitman, Jia, and Gitman}]{openmath}
Shubham Toshniwal, Ivan Moshkov, Sean Narenthiran, Daria Gitman, Fei Jia, and Igor Gitman. 2024.
\newblock Openmathinstruct-1: A 1.8 million math instruction tuning dataset.
\newblock \emph{arXiv preprint arXiv:2402.10176}.

\bibitem[{Wang et~al.(2024)Wang, Li, Shao, Xu, Dai, Li, Chen, Wu, and Sui}]{shepherd}
Peiyi Wang, Lei Li, Zhihong Shao, Runxin Xu, Damai Dai, Yifei Li, Deli Chen, Yu~Wu, and Zhifang Sui. 2024.
\newblock Math-shepherd: Verify and reinforce llms step-by-step without human annotations.
\newblock In \emph{Proceedings of the 62nd Annual Meeting of the Association for Computational Linguistics (Volume 1: Long Papers)}, pages 9426--9439.

\bibitem[{Wei et~al.(2022)Wei, Wang, Schuurmans, Bosma, Xia, Chi, Le, Zhou et~al.}]{cot}
Jason Wei, Xuezhi Wang, Dale Schuurmans, Maarten Bosma, Fei Xia, Ed~Chi, Quoc~V Le, Denny Zhou, et~al. 2022.
\newblock Chain-of-thought prompting elicits reasoning in large language models.
\newblock \emph{Advances in neural information processing systems}, 35:24824--24837.

\bibitem[{Yu et~al.(2023)Yu, Jiang, Shi, Yu, Liu, Zhang, Kwok, Li, Weller, and Liu}]{metamath}
Longhui Yu, Weisen Jiang, Han Shi, Jincheng Yu, Zhengying Liu, Yu~Zhang, James~T Kwok, Zhenguo Li, Adrian Weller, and Weiyang Liu. 2023.
\newblock Metamath: Bootstrap your own mathematical questions for large language models.
\newblock \emph{arXiv preprint arXiv:2309.12284}.

\bibitem[{Zhang et~al.(2024)Zhang, Wu, Lei, Che, Li, Xie, Huang, Zhang, Pavone, Li et~al.}]{berry}
Di~Zhang, Jianbo Wu, Jingdi Lei, Tong Che, Jiatong Li, Tong Xie, Xiaoshui Huang, Shufei Zhang, Marco Pavone, Yuqiang Li, et~al. 2024.
\newblock Llama-berry: Pairwise optimization for o1-like olympiad-level mathematical reasoning.
\newblock \emph{arXiv preprint arXiv:2410.02884}.

\end{thebibliography}



\end{document}